# Harnessing Large Vision and Language Models in Agriculture: A Review

Hongyan Zhu[a, b, *], Shuai Qin[a, b], Min Su[a, b], Chengzhi Lin[a, b], Anjie Li[a, b], Junfeng Gao[c, *]

[a] Guangxi Key Laboratory of Brain-inspired Computing and Intelligent Chips, School of Electronic and Information Engineering, Guangxi Normal University, Guilin, China.

[b] Key Laboratory of Integrated Circuits and Microsystems (Guangxi Normal University), Education Department of Guangxi Zhuang Autonomous Region, Guilin, China.

[c] Department of Computer Science, University of Aberdeen, UK.

* Corresponding authors at: Guangxi Key Laboratory of Brain-inspired Computing and Intelligent Chips, School of Electronic and Information Engineering, Guangxi Normal University, Guilin, 541004, China; Department of Computer Science, University of Aberdeen, UK.

Email address: hyzhu-zju@foxmail.com (Hongyan Zhu); junfeng.gao@abdn.ac.uk (Junfeng Gao).

**Abstract**

Large models can play important roles in many domains. Agriculture is another key factor affecting the lives of people around the world. It provides food, fabric, and coal for humanity. However, facing many challenges such as pests and diseases, soil degradation, global warming, and food security, how to steadily increase the yield in the agricultural sector is a problem that humans still need to solve. Large models can help farmers improve production efficiency and harvest by detecting a series of agricultural production tasks such as pests and diseases, soil quality, and seed quality. It can also help farmers make wise decisions through a variety of information, such as images, text, etc. Herein, we delve into the potential applications of large models in agriculture, from large language model (LLM) and large vision model (LVM) to large vision-language models (LVLM). After gaining a deeper understanding of multimodal large language models (MLLM), it can be recognized that problems such as agricultural image processing, agricultural question answering systems, and agricultural machine automation can all be solved by large models. Large models have great potential in the field of agriculture. We outline the current applications of agricultural large models, and aims to emphasize the importance of large models in the domain of agriculture. In the end, we envisage a future in which famers use MLLM to accomplish many tasks in agriculture, which can greatly improve agricultural production efficiency and yield.

***Keywords.*** large model; agriculture; computer vision (CV); multimodal large language model (MLLM).

# 1. Introduction

*1.1. The challenges facing the agricultural domain*

The significance of agriculture in the global economy is increasing steadily, and there is growing awareness regarding its sustainability. Ahirwar et al. believe that it is necessary to increase global agricultural food production by a minimum of 70% to meet the needs of the increasing world population [1]. Unfortunately, there are many factors in agriculture that make it difficult to steadily increase grain production, including 1) crop diseases caused by pathogens such as bacteria, fungi, and viruses; 2) unscreened low-quality seeds lead to unhealthy crop growth, decreased yield, and susceptibility to crop diseases; 3) many agricultural tasks are inefficient, including weeding, planting, watering, and harvesting crops. Agricultural production is facing enormous economic and production losses. For crop diseases, traditional detection methods like polymerase chain reactions on the basis of unique deoxyribonucleic acid sequences of pathogens, enzyme-linked immunosorbent assays on the basis of pathogens proteins and hyperspectral imaging, are constrained by their operational complexity and the requirement for bulky instruments [2]. For selecting high-quality seeds, quality assurance programs employ various ways to attest seed quality attributes, including germination and vigour tests [3]. But these methods have limitations in terms of time overhead, subjectivity, and the destructive nature of assessing seed quality [4]. For a general tasks in agriculture, the use of pesticides for weed control may have negative impacts on the environment, and Phytotoxicity reactions can lead to diminished crop quality and reduced yields [5]. And the traditional solutions to these tasks are also inefficient due to these manned implements are dreadfully slow. Therefore, it is necessary to develop a method that is fast, simple, and easy for people engaged in agricultural work to use address the problems of the traditional methods mentioned above.

On the other hand, driven by growing health consciousness, the public has long been worried about the safety and quality of food, which is linked to agricultural products. Reducing food losses and improving food safety rely significantly on the continuous monitoring of crop quality, especially the inspection of diseases during crop growth stage [6]. As an efficient analytical means, large model, has found extensive application in the agricultural sector [7] [8] [9]. Large model has shown excellent performance in analysing agricultural data, pest and disease management, precision agriculture, and more. However, it still faces many problems such as difficulty in obtaining agricultural data [10], low model training efficiency, distribution shift[11], and plant blindness [12].

We aim to offer a comprehensive analysis of large model, starting with a systematic summary of the history of large model (LLM and LVM), large model in other fields, the importance of large model for agriculture. Subsequently, introduce many applications of large model in agriculture. Moreover, due to the fact that large models are a relatively new technological means, we outline some solutions from their ethical and responsibility aspects. Finally, summarize the current challenges and future directions of large models, and draw conclusions on the effectiveness of its implementation in the agricultural domain.

*1.2. The history of LLM and LVM*

Artificial intelligence (AI), whose main purpose is to establish systems that learn and think like human [13], just like human language and visual abilities. At present, research on large models is also focused on the natural language processing (NLP) and computer vision (CV). Next, LLMs and LVMs will be introduced in detail. LLM is a model based on NLP, and we can divide the development of it into four stages:

- Statistical language models (SLM). SLMs use traditional statistical techniques such as n-gram and some language rules to learn the probability distribution associated with words. It is generally believed that the amount of data, and the ability of a given estimation algorithm to accommodate large number of training are very significant in providing a solution that competes successfully with the entrenched n-gram language models [14]. SLMs are currently widely used in the field of NLP, such as Raychev et al. designed a simple and scalable static analysis that uses SLMs to complete incorrect code [15]. However, n-gram models have three drawbacks. Firstly, as n increases, the more parameters need to be calculated and counted, and the more memory space it occupies. Markov assumption can be used to limit the size of n [16]. Secondly, n-grams models cannot share information from vocabulary or prefixes with the same semantics. Word embedding can be used to shift character representation to vector representation [17]. Thirdly, data sparsity. Data smoothing, backoff [18] and interpolation can be used to solve this problem. In addition, neural network models can better handle the problem of data sparsity.

- Neural language models (NLM). NLMs [19] [20] [21] use different types of neural networks to model language, and compared to SLMs, NLMs are more effective. To solve the problem of data sparsity in n-gram models, feedforward neural networks and recurrent neural networks (RNN) were used in continuous space language modelling, which can enable the model to automatically learn features and continuous representations. The first feedforward neural network language model (FFNNLM) was proposed by Bengio et al. in 2003, which overcomes the curse of dimensionality by learning distributed representations of a word [19]. Subsequently, Mikolov et al. suggested the RNN language model (RNNLM), which can make predictions using limited context [20]. However, during the training process of RNNLM, the gradient of parameters may disappear or explode, leading to slower training speed or infinite parameter values, making it difficult for the model to achieve long-term dependence. Sundermeyer et al. applied long short-term memory recurrent neural networks to language models in 2012 and proposed LSTM-RNNLM [21]. Three gate structures (including input, output, and forget gates) had been added to the LSTM memory unit to control information flow, which solved the problem of long-term dependence in language models learning.
- Pre-trained language models (PLM). PLMs can be divided into two paradigms: feature-based and fine-tuning. Feature-based treats pre-training as a feature extraction process, trains model parameters on large-scale corpora, and encodes them as fixed features to downstream models for collective tasks. A typical example is ElMo, a pre-training bidirectional LSTM (BiLSTM) proposed by Peters et al. [22]. Due to LSTM modelling sentences, it can solely consider the contextual information preceding the current sentence and fails to capture subsequent contextual information. And BiLSTM uses reverse networks, which can concurrently consider contextual information before and after, thus better processing sequential data. Fine-tuning transfers the parameters of the entire model to downstream tasks, which is the current mainstream paradigm and has better flexibility compared to feature-based. The representative models of fine-tuning are bidirectional encoder representations from transformers (BERT) [23] and generative pre-trained transformer (GPT). In 2017, Google's research team proposed Transformer [24], a model that uses self-attention mechanism, while OpenAI proposed GPT based on the architecture of Transformer. GPT achieved almost perfect training results by pre-training on large-scale text datasets and fine-tuning parameters. BERT was proposed by pre-training bidirectional language models with especially designed pre-training tasks on vast unlabelled corpora. These pre-trained context-aware word representations demonstrate high efficacy as general-purpose semantic features, significantly enhancing the performance of NLP tasks [25]. Furthermore, due to the significant acceleration of model training by Transformer, it has gradually become the fundamental architecture for LLMs.

- Large language models (LLM). LLM is a language models that contains billions (or more) of parameters. Large models possess abilities that small models do not possess, which is known as the emergence abilities of LLMs [26]. This is also one of the most obvious distinguishing features between LLMs and PLMs. OpenAI researchers discovered that larger models will continue to exhibit better, and will also be much more sample efficient than before [27]. Many current studies have trained large-sized PLMs and found that compared to smaller PLMs, large-sized PLMs exhibit different behaviours and exhibit astonishing abilities in solving a range of complex tasks [25]. This is the emergence abilities of LLMs, as mentioned earlier. For instance, the context learning ability of GPT-3 can generate the expectant output of test examples by completing word sequences of input text, without the need for extra training or gradient updates, which GPT-2 cannot achieve. Therefore, the research community refers to these large-sized PLMs with additional capabilities as LLMs [28] [29] [30].

LVM is a model associated with CV. The research on vision models initially focused on shallow image feature extraction algorithms, including scale-invariant feature transform, histogram of oriented gradient, and other methods, but had significant limitations. In 2012, AlexNet [31] achieved a breakthrough success in ImageNet large scale visual recognition challenge, sparking a wave of convolutional neural networks (CNN) for vision models. With the development of deep learning, deep residual networks including VGGNet [32], GoogLeNet [33], and ResNet [34] were successively proposed, which improved the performance of image classification, object detection, semantic segmentation, etc. The boom of the Internet also enabled large-scale image datasets to be used for training vision models. Faster R-CNN [35], YOLO [36], Mask R-CNN [37] emerged one after another. In recent years, Transformer has been applied in the domain of LVM, and vision transformer (ViT) [38] and DALL-E [39] have appeared in front of the public. These models use self-attention mechanism, combined with generative adversarial network, to demonstrate strong capabilities in image classification and generation tasks.

In addition to the LLM and LVM introduced above, multimodal large language models (MLLM) are also a research focus in the domain of AI. LLMs perform well in text-based tasks, but they are hard to understand and handle other data types. LVMs perform well in the field of CV, but there is limited information on the analysis results, which has certain limitations for users. MLLMs [40] integrate multiple data types, such as images, text, language, audio, and more. It not only possesses the advantages of LLMs and LVMs, but also address the limitations of LLMs and LVMs by integrating multiple modalities, enabling a more comprehensive understanding of various data. It can be said that the developments in MLLMs have set up new avenues for AI, which make binary machines to understand and then process various data types [40].

*1.3. The current developed large models*

Currently, many industry professionals have found that large models can bring breakthroughs to their industries. In order to create a large model that is suitable for the industry and can complete some professional tasks (Table 1), different industries have begun to consider invested manpower, material resources, and financial resources in succession.

*1.4. The current large models in other domains*

As shown in Table 1, the current large models are mainly LLMs and MLLMs, with LVMs accounting for a minority. Many LLMs are designed to develop chatbots (BLOOM [41], PaLM2 [42], ERNIE Bot) or complete NLP tasks, including text classification, machine translation, and sentiment analysis (OPT [43]). Some researchers are not satisfied with NLP tasks, so they have added visual ability to enable the model to answer questions based on images (Minipt-4 [44]) , this type of model can be called large vision-language model (LVLM).

Although LVLM satisfies some functions and takes large models a big step towards artificial general intelligence (AGI), it is not enough to achieve the goal that machines can emulate human thinking and carry out a wide range of general tasks through transfer learning and diverse other modalities without achieving the multimodality of the model [45]. Some large models have implemented multimodality, enabling them to analyze different types of information (GPT-4 [46], LLaMA [47], Gemini [48], ImageBind [49]) and interact with users.

**Table 1**

The currently popular large models.

| LLM | Release date | Open source | Models | Information | References |
|---|---|---|---|---|---|
| Chinchilla | March 29th, 2022 | No | LLM | Training a large language model with optimal computational utilization. | [30] |
| OPT | May 2rd, 2022 | Yes | LLM | The full name of OPT is open pre trained transformer language models, which means "open pre trained transformer language models". | [43] |
| BLOOM | November 9th, 2022 | Yes | LLM | A decoder only model based on Transformer architecture. | [41] |
| Minigpt-4 | April 20th, 2023 | Yes | LLM | This model is capable of understanding images and text and responding to user instructions. | [44] |
| LLaMA-Adapter-V2 | August 28th, 2023 | Yes | LLM | A Parameter Efficient Visual Instruction Model. | [47] |
| PMC-LLaMA | Aprill 27th, 2023 | Yes | LLM | Inject medical knowledge into existing LLM using 4.8 million biomedical academic papers. | [50] |
| PaLM-E | March 6th, 2023 | No | LLM | PaLM-E is a large language model with only a decoder. | [51] |
| PaLM2 | May 11st, 2023 | No | LLM | PaLM2 is a neural network-based language model that is considered one of the most advanced language models currently available. | [42] |
| ERNIE Bot 4.0 | October 17th, 2023 | No | LLM | ERNIE Bot is a new generation of Baidu's big language model for knowledge enhancement. | / |
| Qwen-7B | August 3rd, 2023 | Yes | LLM | A super large language model launched by Alibaba Cloud. | [52] |
| IFLYTEK SPARK | May 6th, 2023 | No | LLM | IFLYTEK SPARK is a new generation of cognitive intelligence model with Chinese as its core. | / |
| BloombergGPT | March 30th, 2023 | No | LLM | A LLM for the financial field. | [53] |
| OCEANGPT | October 3rd, 2023 | Yes | LLM | A LLM for ocean science tasks. | [54] |
| InternImage | April 17th, 2023 | Yes | LVM | A visual large model based on deformable convolution. | [55] |

**Table 1** (*continued*).

| LLM | Release date | Open source | Models | Information | References |
| --- | --- | --- | --- | --- | --- |
| PanguCVLM | April 25th, 2021 | No | LVM | PanguCVLM is a model that utilizes a large model network to simulate and automate human visual processes. | / |
| LLaVA | April 17th, 2023 | Yes | MLLM | LLaVA, a new large multimodal model. | [56] |
| Instruct BLIP | June 15th, 2023 | Yes | MLLM | Instruct BLIP model achieves state-of-the-art zero sample performance on a wide range of visual language tasks. | [57] |
| Visual ChatGPT | March 8th, 2023 | Yes | MLLM | The proposal of Visual ChatGPT opened the door to the connection between ChatGPT and VFM, enabling ChatGPT to handle complex visual tasks. | [58] |
| mPLUG-Owl | April 27th, 2023 | Yes | MLLM | A multimodal dialogue generation model similar to miniGPT-4 and LLaVA, with image viewing and chat functionality. | [59] |
| VisualGLM-6B | May 17th, 2023 | Yes | MLLM | VisualGLM-6B is an open-source multimodal dialogue language model that supports images, Chinese, and English. | [60] |
| ImageBind | May 9th, 2023 | Yes | MLLM | ImageBind is the first artificial intelligence model that can bind information from six modes. | [49] |
| MultiModal-GPT | May 8th, 2023 | Yes | MLLM | MultiModal GPT can follow various human instructions, such as generating detailed instructions, calculating the number of objects of interest, and answering general questions from users. | [61] |
| GPT-4 | March 14th, 2023 | No | MLLM | GPT-4 is the latest pre trained model launched by OpenAI, aimed at improving the performance of machine learning models using NLP technology. | [46] |
| Skywork | April 17th, 2023 | No | MLLM | Skywork is a series of large models developed by the Kunlun · Skywork team. | [62] |
| Gemini | December 6th, 2023 | No | MLLM | Gemini is an artificial intelligence multimodal large language model launched by Google DeepMind. | [48] |
| Sora | February 15th, 2024 | No | MLLM | Sora is an AI model that can create realistic and imaginative scenes from text instructions. | [63] |

However, many current models are generic models and their training datasets are too broad, they cannot provide a satisfactory answer to knowledge in certain professional fields. As Goertzel believed, for a system to be considered AGI, it is not necessary for it to have infinite generality, adaptability, or flexibility [64]. Therefore, some researchers have optimized and adjusted existing large models and have released some large models specifically for a single field. BloombergGPT can be used in the financial field, showcasing remarkable performance on general LLM benchmarks and surpassing comparable models on financial tasks [53]. The meteorological model in panguLM developed by Huawei can provide predictions of variables such as gravity potential, humidity, wind speed, temperature, and pressure within 1 hour to 7 days. Embedding PaLM-E into robots can achieve multiple specific tasks, like visual question answering, sequential robotic manipulation planning, and captioning [51]. OCEANGPT is an expert in various marine science tasks [54]. It exhibits not only a higher level of knowledge expertise for oceans science tasks but also acquires preliminary embodied intelligence capabilities in ocean engineering. PMC-LLaMA represents the pioneering open-source medical specific language model that demonstrates exceptional performance on diverse medical benchmarks, outperforming ChatGPT and LLaMA-2 with significantly fewer parameters [50]. Obviously, large models can be applied in many fields. Naturally, it also includes agriculture.

*1.5. The importance of large models in the agricultural domain*

In the past few decades, the advancement of agricultural technology has significantly improved global agricultural production efficiency. According to the forecast released by the food and agriculture organization (FAO) of the United Nations, the global grain production in 2023 was 2.84 billion tons, nearly twice that of the early 20th century. Although global agricultural production efficiency is high, the world population is also constantly growing. Continuously improving agricultural production efficiency is the lifeline of economic development and the foundation for ensuring human food, clothing, and survival needs. Hence, how to make agricultural practices advance is a crucial issue. Currently, many challenges faced by agriculture can be addressed using large models:

- Crop pests and diseases are one of the major problems in agricultural production. Most farmers blindly spray pesticides to control the occurrence of diseases and pests, which has caused many problems including environmental pollution and food safety [65]. The traditional identification of crop pests and diseases primarily relies on the expertise and experience of farmers and agricultural experts. Although these experiences are reliable, it is inevitable that misjudgements may occur. Moreover, some large plantations have more crops, requiring more manpower and time, making it easy to miss the appropriate time to rescue diseased crops. The use of large models can lower the threshold for detecting pests and diseases, and greatly improve the efficiency of identification. For example, farmers can use a large model targeting diseases and pests, take photos of crops suspected of having diseases, and upload them to the large model, which will analyse the results of pests and diseases.
- Weeding issues. Weeds are analogous in colour to crops, and their growth also needs sunlight and water. This leads to weeds and crops competing for survival space and nutrients. Excessive weed density in the field can significantly impact the yield and quality of crops. Hence, weeding is an inevitable and vital link in agricultural yield. Farmers can use agricultural robots embedded in large models to identify and trim weeds [51].
- The quality of seeds determines the survival rate and yield of crops. When high vitality seeds are planted in the field, due to their strong stress resistance, they have a high germination rate, fast and neat emergence, and can produce high-yield and high-quality crops. Seeds with low vitality exhibit weak stress resistance and are prone to rot and die, ultimately leading to a decrease in crop yield. Using large models to accurately identify high vitality seeds can greatly improve crop yield and quality.

- In addition to identifying seed quality, grading mature crops is also important. It involves strict screening of crops before entering the market, removing decaying crops, and classifying them based on their quality. Large models can detect crop photos and efficiently grade crops, greatly reducing the time required for crops to enter the market stage.

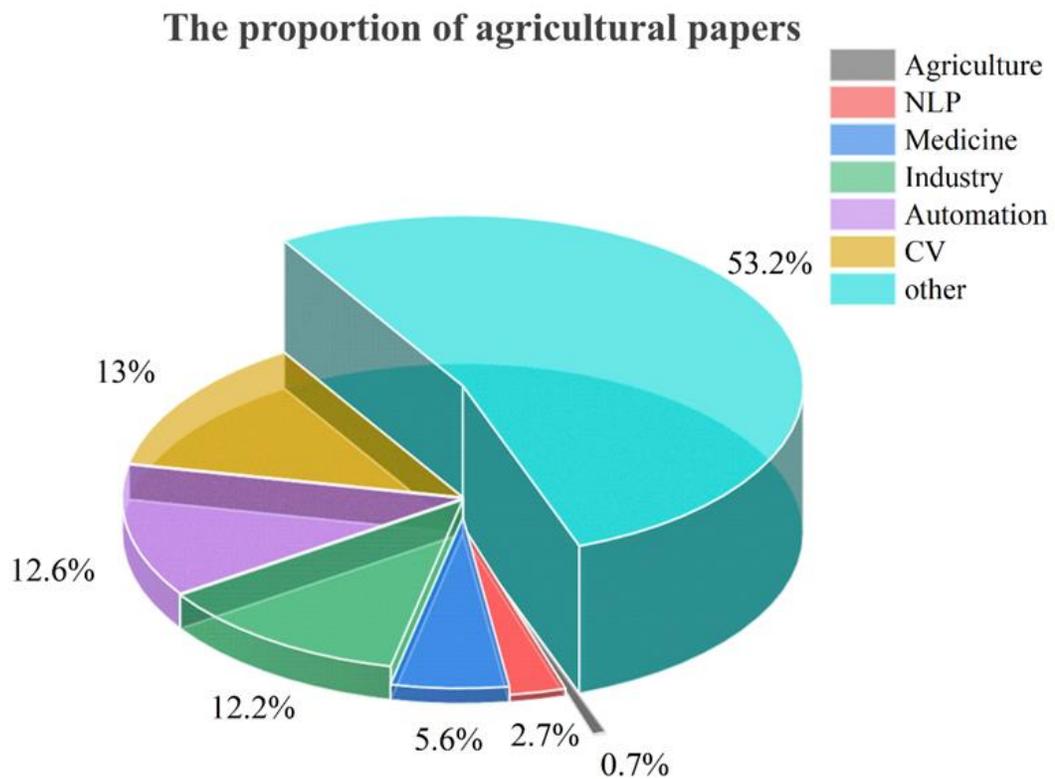

**Fig 1.** The proportion of arxiv papers on agriculture in AI. A kaggle dataset from arxiv (CS. AI domain, a total of 3496 papers published after 2019 were selected from 10000 papers) was subjected to keyword screening in different fields, targeting agriculture (24 papers), NLP (94 papers), medicine (195 papers), industry (425 papers), automation (442 papers), and CV (455 papers). 1862 unselected papers remaining.

Large models play such a crucial role in agriculture, but as shown in Fig 1, the proportion of agricultural papers in the field of AI is currently too small compared to other fields, making it difficult to receive people's attention. Based on the title and the above content, our research objective is already very clear. That is to study what are the large models that can be used in agriculture, and how to use them. The following chapters will gradually introduce the application prospects and practical applications of LLM and LVM in agriculture, finally revealing the importance of large models in the sector of agriculture.

## 2. Large Language Models in Agricultural Applications

Numerous agricultural tasks require intricate reasoning. For example, when presented with an image of a soybean field, agricultural scientists or farmers rely on large models to undertake several key steps. Firstly, the large model must identify any abnormal symptoms evident in the soybean leaves, such as water stains. Subsequently, it must ascertain the name of the specific problem that troubles plants, such as bacterial wilt disease. Next, the model needs to determine the underlying cause of the disease, such as pseudorabies. Finally, it must develop an appropriate treatment strategy, such as applying a bactericide spray.

Many question answering (QA) and dialogue systems are designed to address this type of reasoning problem [66] [67] [68]. For instance, a chatbot based on a RNN is specifically designed to handle questions related to soil testing, plant protection, and nutrient management [66]. Although, these QA and dialogue systems and chatbots are capable of answering most inquiries without the need for human interaction and with excellent accuracy, they have limited capabilities for complex problems by reason of their small model size as well as of inadequate training data. Therefore, the agricultural domain requires large models to promote the development of QA and dialogue systems and chatbots.

*2.1. The role of LLM in processing and generating agricultural data, providing insights and decision-making support*

LLM can play many roles in the agricultural domain, such as processing and generating agricultural data, providing insights into agricultural production work, and supporting agricultural decision-making for farmers.

2.1.1. LLMs processing and generating agricultural data

**Information extraction**. LLMs can extract structured information from unstructured agricultural text data. First, the text is divided into individual tokens and LLMs represent each token as a numerical vector called a word embedding. Then, LLMs analyse the surrounding context of each token to understand its meaning within the sentence or document, and identify and categorize named entities within the text, like names of individuals, locations, organizations, or specific agricultural terms. Finally, LLMs employ techniques like information extraction to identify and extract structured information from unstructured text (Involve identifying relationships between entities, extracting key facts, or populating knowledge graphs). LLMs extract information from data using a process known as NLP. Peng et al. used LLM (Not related to the agricultural domain) to automatically extract entities and attributes from unlabelled agricultural data and transform them into structured data [69]. Information extraction is beneficial for LLM to better understand the overall meaning of the text.

**Agricultural data generation**. Generative AI models are actually a multimodal LLMs (This content will be detailed in 4.1). An obstacle encountered when applying specialized CV algorithms to agricultural vision data is the insufficient availability of training data and labels [70] [37]. In addition, collecting data that encompasses the wide range of variations caused by season and weather changes is exceedingly challenging. Acquiring high-quality data requires a lot of time, and labelling them is even more costly [71]. To address these challenges, one approach is to fine-tune multimodal generative LLMs on the target agricultural data domain. This allows the models to generate massive training data and labels, thereby constructing an augmented training set that closely resembles the distribution of the original data [72]. Besides, text-based generation models can generate images [73] and videos [74] of specific scenes based on text descriptions, thereby supplementing training datasets that may lack certain visual content. These models can also be employed to generate diverse variations of the initial data for some characteristics: multimodal generative LLMs can transform the time of the image from daytime to nighttime [75] or alter the weather conditions from rainy to sunny. This helps in expanding the training data and improving the performance of downstream models.

2.1.2. LLMs provide insights

LLMs possess the capability to analyse textual data and uncover trends in agricultural practices, market conditions, consumer preferences, and policy developments. Through analysis of agricultural text data from sources such as news articles, reports, and social media, these models can offer valuable insights into market dynamics and pricing trends [76]. This provides support for farmers to understand domains outside of agriculture. Many researchers believe that the integration of LLMs into different stages of designment and development for agricultural applications is also experiencing a noticeable rise [8] [77]. In [8] study, Stella et al. incorporated LLM into the design phase of robotic systems. They specifically focused on designing an optimized robotic gripper for tomato picking and outlined the step-by-step process. In the initial ideation phase, they leveraged LLMs like ChatGPT [46] to gain insights into the possible challenges and opportunities associated with the task. Building upon this knowledge, they identified the most promising and captivating pathways, engaging in ongoing discussions with the LLM to refine and narrow down the design possibilities. Throughout this process, the human collaborator harnesses the expansive knowledge of the LLM to tap into insights transcend their individual expertise. In the following stage of the design process, which emphasizes technical aspects, the broad directions derived from the collaboration need to be transformed into a real, completely functional robot. Although LLMs do not provide comprehensive technical support, they can offer their own insights on whether the technology is feasible, helping researchers reduce the risk of failure.

Presently, LLMs lack the ability to generate comprehensive CAD models, evaluate code, or independently fabricate robots. Nevertheless, advancements in LLM research suggest that these algorithms can offer significant assistance in executing software [78], mathematical reasoning [79], and even in the generation of shapes [80]. Lu et al. specifically focused on the utilization of LLMs for organizing unstructured metadata, facilitating the conversion of metadata between different formats, and discovering potential errors in the data collection process [77]. They also envisioned the next generation of LLMs as remarkably potent tools for data visualization [46], and anticipated that these advanced models will provide invaluable support to researchers, enabling them to extract meaningful insights from extensive volumes of phenotypic data.

Although LLMs provide insights can indirectly help farmers solve a small number of agricultural tasks, it's important to note that their insights should be used in conjunction with human judgment and domain expertise. That is to say, the insights provided by LLMs cannot be separated from human experience.

2.1.3. LLMs provide decision-making support for farmers

According to a recent study, ChatGPT demonstrates the ability to comprehend natural language requests, extract valuable textual and visual information, select appropriate language and vision tasks, and effectively communicate the results to humans [81]. Shen et al. proposed a system named HuggingGPT to solve AI tasks. The system connects LLM with AI models through language interface, and these AI models are derived from HuggingFace. This indicates that LLMs can serve as the core of decision-making, allowing this core to control other AI models, thus enabling the entire system to address the tasks proposed by farmers.

As the core of decision-making, LLM can be applied to agriculture to help solve the tasks proposed by farmers [81].

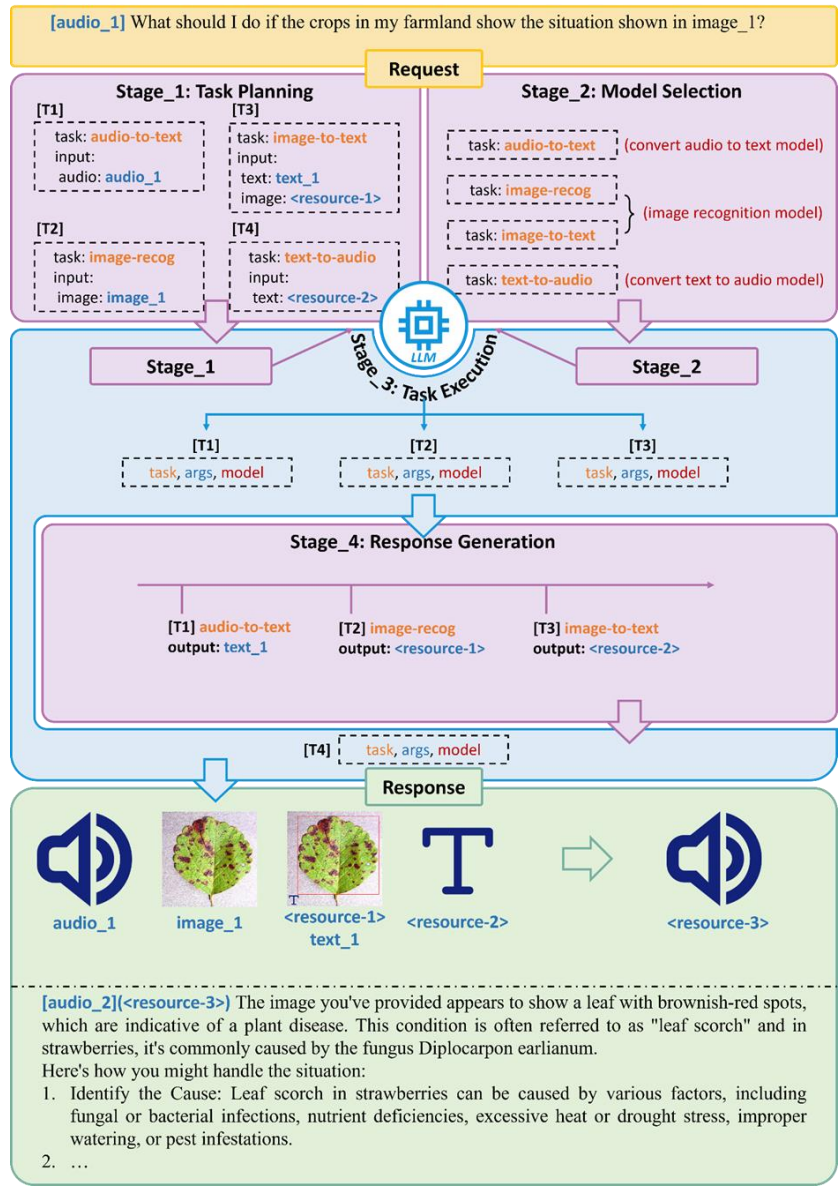

**Fig 2.** LLMs provide decision-making support.

LLMs have the potential to function as controllers, overseeing and managing the operations of existing AI models to address complex AI tasks. As shown in Fig 2, A farmer with a low level of education used audio to consult the system and sent an image. When receiving a task request, LLM first divides the total task into subtasks and selects the appropriate AI model based on the needs of each subtask. For example, converting farmers' audio into text requires the use of an audio to text model (Amazon transcribe, Whisper [82]); It is also necessary to recognize the sent image and integrate the text obtained from the audio conversion in the previous step to obtain a text-response (vit-gpt2); Considering that the farmer is illiterate, it is necessary to further convert text-response into audio and ultimately obtain the audio-response (Fastspeech [83] [84]). Although LLM does not play a role in solving problems throughout the entire system, as a "conductor", it can coordinate various AI models to complete subtasks, thereby gradually solving complex tasks and playing a core role in decision-making support.

*2.2. Few-shot learning of LLM in the agricultural domain*

In the previous section, it was mentioned that LLM plays a significant role in processing and generating agricultural data, providing insights and decision support. But in the case of fully-supervised learning, this often requires a large number of labelled samples to train in order to obtain a good model. Unlike many current models, human beings possess the remarkable capacity to derive new knowledge even in situations where they have limited or zero experiences. To narrow the gap between human beings and large models, researchers have proposed the concept of few-shot learning (FSL) [85]. FSL only requires meta knowledge (Prior knowledge) to infer new knowledge, that is, in the case of insufficient samples, FSL can also obtain remarkable generalization ability based on limited samples. It is very useful for agricultural domain where data collection and labelling are difficult and expensive.

FSL has already shown superior performance in the NLP domain [86] [87] [88], it can be categorized into three distinct types: few-shot, one-shot, and zero-shot. Few-shot applies to 2 to 5 samples per class, one-shot applies to one sample per class, and zero-shot classifies invisible classes without samples. In a study conducted by Tom B. Brown et al., GPT-3 was trained and evaluated in few-shot setting. The results indicated that GPT-3 demonstrates strong performance in diverse NLP tasks, such as translation, QA, and cloze [89].

Although FSL performs well in image classification, object detection, and object tracking, its generalizability is very limited compared to fully-supervised learning. Consider two examples of classifying fruits in an image and detecting fruits in an image. While both tasks involve recognizing fruits, they cannot be directly converted into one another. Consequently, each task necessitates a dedicated model specifically trained for that particular objective, which greatly reduces the generalizability of the model. The main problem with training models in the agricultural domain is the lack of labelled data, which leads to poor generalization of models trained by fully supervised learning. For this reason, researchers are trying to use FSL to reduce the demand for particular crop data. Surprisingly, LLMs not only has excellent performance in generalizability, but also demonstrate an innate talent for few-shot learning [89].

LLMs have demonstrated remarkable abilities across diverse domains, exhibiting the capability of zero-shot generalization without the requirement for fine-tuning specific to each task [90] [91]. Even so, these models are primarily limited to processing text-based data. MLLMs may be able to overcome these limitations. It is worth noting that many current research directions for large models still only focus on image, neglecting audio, video, and other aspects. As the most advanced large model of OpenAI, GPT-4 only supports input of text, audio and images [46] [92]. Large models with visual ability show their versatility in different domains [55]. Because the images used in training models are from the Internet, these images have a large gap with the real agricultural images, which leads to these models cannot be well applied to the agricultural domain. As a consequence, agriculture researchers can only rely on FSL to train a large model suitable for agriculture, unless there are suitable agricultural images to train large vision-language models (LVLM).

## 3. Large Vision Models in Agricultural Applications

The public often confuse LVLM and LVM. LVLM refers to LLM with visual ability. LVM is a purely visual large model that does not making use of any linguistic data, and only uses image data for training and inference [93]. The goal of LVM is to learn universal visual knowledge and adapt to different visual tasks and scenes.

The current use of large models in agriculture is mainly focused on CV. Using models to analyze diseases, pests, weeds, seeds, mature crops, and other aspects involves the use of LVMs, among which the main problem is still the problem of diseases and pests. The traditional methods for detecting crop pests and diseases mainly rely on special methods such as serology and molecular biology-based technical means, in addition to artificial visual evaluation. Although these methods can accurately determine pests and diseases to a certain extent, they often require a lot of time and money. And some methods of sampling crops often lead to crop damage, which goes against the original intention of diagnosing diseases and pests to protect crops. Therefore, image processing and analysis is an important task for large models in the field of agriculture, and another important task is to embed LVMs into robots to solve some agricultural problems (Weeding, pruning branches, harvesting, etc.) and achieve automated agriculture.

*3.1. Image processing and analysis*

Using a LVM to judge crop related information can not only greatly improve the time required for judgment, but also indirectly reduce the damage caused to crops. Moreover, after crops are invaded by pests and diseases, their color, texture, spectral characteristics will undergo certain changes, all of which are related to CV.

At present, there are four types of methods for obtaining crop image information: 1) ordinary channels, taking photos to obtain images; 2) obtaining remote sensing images through agricultural machinery near the ground; 3) obtaining remote sensing images through aircraft monitoring platforms [94]; 4) obtaining remote sensing images through satellites [95]. Remote sensing can provide large-scale land use and land cover information. By analyzing satellite images or high-altitude images, various surface information can be identified, such as surface conditions, soil moisture, vegetation coverage, and crop growth status [96]. Classifying and segmenting from limited examples obtained from remote sensing is a significant challenge. Regarding this, Wu et al. put forward GenCo (a generator-based two-stage approach) for few-shot classification and segmentation on remote sensing and earth observation data [97]. Their approach presents an alternative solution for addressing the labeling challenges encountered in the domains of remote sensing and agriculture. Spectral data can provide rich insights into the composition of observed objects and materials, especially in remote sensing applications. The challenges faced in processing spectral data in agriculture include: 1) effectively processing and utilizing vast amounts of remote sensing spectral big data derived from various sources; 2) deriving significant knowledge representations from intricate spatial-spectral mixed information; 3) addressing the spectral degradation in the modeling of neighboring spectral relevance. Hong et al.'s SpectralGPT empowers intelligent processing of spectral remote sensing big data, and this LVM has also demonstrated its excellent spectral reconstruction capabilities in agriculture [98]. Due to multispectral imaging (MSI) and hyperspectral imaging (HSI) make it possible to monitor crop health in the field. The integration of remotely sensed multisource data, such as HSI and LiDAR (Light detection and ranging), enables the monitoring of changes occurring in different parts of a plant [99]. By using a large visual model to analyze these spectral data, the obtained crop health information can help farmers quickly and accurately identify diseases and treat them, reducing the loss of crop yield.

The use of LVMs for image recognition and predictive analysis of crop information is often more effective than traditional machine learning algorithms. When farmers need to obtain crop information, four types of image acquisition methods can be used to obtain crop image information (Fig 3). Then, the image information is processed through image recognition (Divided into four tasks: image classification, object detection, semantic segmentation, instance segmentation), and the identified results need to be further predictive analytics by the model (LVLM) to obtain crop information that farmers can understand.

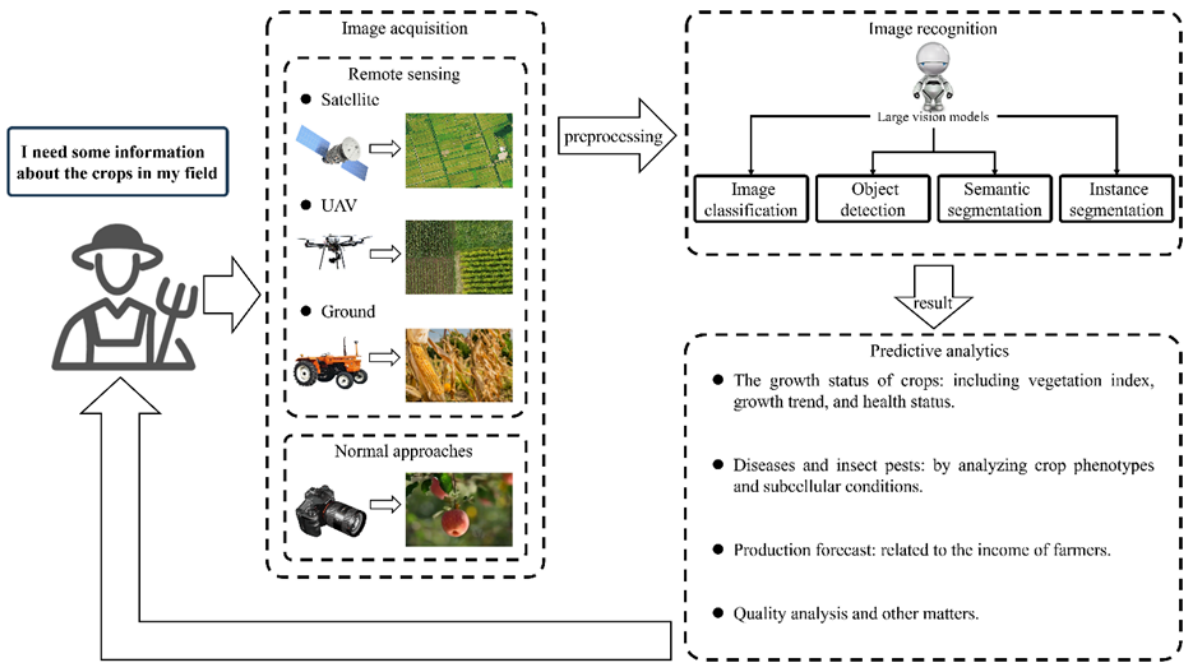

**Fig 3.** Farmers can obtain crop information through the process of image acquisition, image recognition, predictive analytics.

In addition to obtaining information by analyzing the phenotypic characteristics of crops, Feng et al. developed an organelle segmentation network (OrgSegNet) [100]. OrgSegNet is capable of accurately capturing the actual sizes of chloroplasts, mitochondria, nuclei, and vacuoles within plant cell, further inspecting plant phenotypic at the subcellular level. They have tested two applications: 1) A thermo-sensitive rice albino leaf mutant was cultivated at cold temperature conditions. In the transmission electron microscope images (TEMs), the albinotic leaves lacked typical chloroplasts, and OrgSegNet failed to identify any chloroplast structures; 2) Young leaf chlorosis 1 (Ylc1). Young leaves of the ylc1 mutant showed lower levels of chlorophyll and lutein compared to corresponding wild type, and its TEM analysis further revealed a noticeable loose arrangement of the thylakoid lamellar structures. It can be imagined that if a large model is used to replace deep learning algorithms, the recognition of subcellular cells may perform better, and the recognition results can be further predictive analytics to obtain information that non plant experts can also understand.

*3.2. Automation and robotics*

A conventional agricultural robot system consists of perception, decision-making, and actuation modules [101]. Their perception module utilizes CV and deep learning to accurately identify crops, soil conditions, and other relevant information [102]. The module of decision-making utilizes this data to automatically provide suitable agricultural management strategies based on factors such as crop growth status and soil quality [103]. The actuator module is responsible for executing specific tasks as determined by the decision-making module [104]. Nevertheless, traditional agricultural robot systems have limitations in processing large volumes of offline data. They lack high-performance data processing and high-quality actual-time control capabilities. This is due to the potential network communication and computing burdens associated with big data processing, causing decreased system performance and heightened costs [105]. Furthermore, they were usually designed for specific crops based on crop type and application requirements. The drawbacks of traditional systems were evident in their inflexible control logic and the absence of intelligence, including automatic decision-making and motion generation [106]. In response to these challenges, it is necessary to use large models to help lift the intellectual features of agricultural robots.

Current LVMs can be used in drones to monitor crops and obtain information on their growth, disease, yield, and other factors [107] [108] [109]. In addition to the above functions, ground machines that used LVMs can also be used for harvesting and classifying crops, as well as detecting pests up close. In [9], a LVM, segment anything model (SAM) [110], uses infrared thermal images for chicken segmentation tasks in a zero-shot means. SAM can be used in agriculture to segment immature fruits on a fruit tree and quickly achieve yield prediction. Yang et al. subsequently proposed the Track Anything Model (TAM) by combining SAM and video [111]. Unfortunately, TAM places more emphasis on maintaining short-term memory rather than long-term memory. Nevertheless, TAM still has great potential in the agricultural field. If its long-term memory ability can be improved, it can monitor early changes in crop diseases and provide early warning to farmers. Embedding LVMs such as SAM and TAM into robots can not only achieve automation in agriculture, but these LVMs themselves can help achieve automation in agricultural robot design. As mentioned in section 2.1.2, Stella et al. used ChatGPT and other LLMs to assist in designing an optimized robotic gripper for tomato picking [8]. It is worth mentioning that the ChatGPT they used at the time was the GPT-3 version, which only supported text input at this time. At present, ChatGPT has been updated to GPT-4 version and has the function of LVLM. Designers of agricultural robots can input text descriptions and sketches into ChatGPT to achieve partial automation of design robots.

Overall, due to the real-time requirements of agricultural robots, the scale of traditional vision models is small, and the recognition results of agricultural images are often unsatisfactory. The emergence of LVM has broken this deadlock, as pre-trained recognition results using large-scale datasets often outperform traditional vision models. Its disadvantage is that it usually requires more computing resources and time to complete the inference and prediction process, resulting in poor real-time performance. However, by optimizing the model architecture, using efficient inference algorithms, and utilizing hardware acceleration techniques, the real-time performance of LVMs can be improved to a certain extent [112].

*3.3. LVLM compared to LVM*

As mentioned at the beginning of this chapter, LVLM and LVM cannot be confused, but LVLM can be considered as an LVM with NLP capabilities. LVLM is a model that combines both visual and textual information to comprehend and generate captions, respond to questions about images, and tackle tasks that necessitate a comprehensive understanding of both vision and language.

Like LVM, research on LVLM is also very early. Mori et al. conducted research on enhancing content-based image retrieval by training a model to forecast nouns and adjectives in text associated with images as early as 1999 [113]. In 2007, Quattoni et al. showcased the potential of learning more data efficient image representations through manifold learning in the weight space of categorizer trained to predict words in image captions [114]. Over a decade ago, Srivastava and Salakhutdinov delved into the domain of deep representation learning by training multimodal Deep Boltzmann Machines using low-level image and text tag features as a foundation

[115]. Until recent years, the CLIP [116] of Radford et al. and ALIGN [117] of Jia et al. have led to further development of LVLM. Compared to LLM with pure text input, LVLM that can combine text and image is the mainstream of large model research today.

In a word, LVLM and LVM differ not only in concept but also in usage. LVLM requires the combination of text and images to infer image information, while LVM relies solely on images for image processing. Due to the lack of textual information, in most cases, LVM is not as effective as LVLM in inferring image content. It is worth mentioning that LVLM belongs to the simplest type of MLLM, and the development of large models is no longer limited to text and images.

## 4. Multimodal Large Language Model and Model Assessment

*4.1. Integration of multimodal models*

MLLM recently has emerged as a prominent research hotspot, which uses powerful LLMs as a core to tackle multimodal tasks [118]. The most common MLLM (Supporting both images and text) is LVLM (Fig 4). In recent years, many researchers have utilized and merged diverse types of data inputs, such as text, images, audio, video [119], sensor data [51], depth information, point cloud [120], and more.

The agricultural community has started exploring the realm of multimodal learning in agricultural applications. By incorporating multimodal learning techniques, the agricultural community seeks to unlock new opportunities for optimizing various agricultural processes and achieving improved outcomes. As an example, Bender et al. have released an open-source multimodal dataset specifically curated for agricultural robotics [121]. This dataset was collected from cauliflower and broccoli fields and aims to foster research endeavours in robotics and machine learning within the agricultural domain. It encompasses a diverse range of data types, including stereo colour, thermal, hyperspectral imagery, as well as essential environmental information such as weather conditions and soil conditions. The availability of this comprehensive dataset uses as a precious resource for advancing the development of innovative solutions in agricultural robotics and machine learning. Cao et al. proposed a novel approach for cucumber disease recognition using a MLLM that incorporates image-text-label information [122]. Their methodology effectively integrated label information from many domains by employing image-text multimodal contrastive learning and image self-supervised contrastive learning. The approach facilitated the measurement of sample distances within the common image-text-label space. The results of the experiment demonstrated the effectiveness of this innovative approach, achieving a recognition accuracy rate of 94.84% on a moderately sized multimodal cucumber disease dataset.

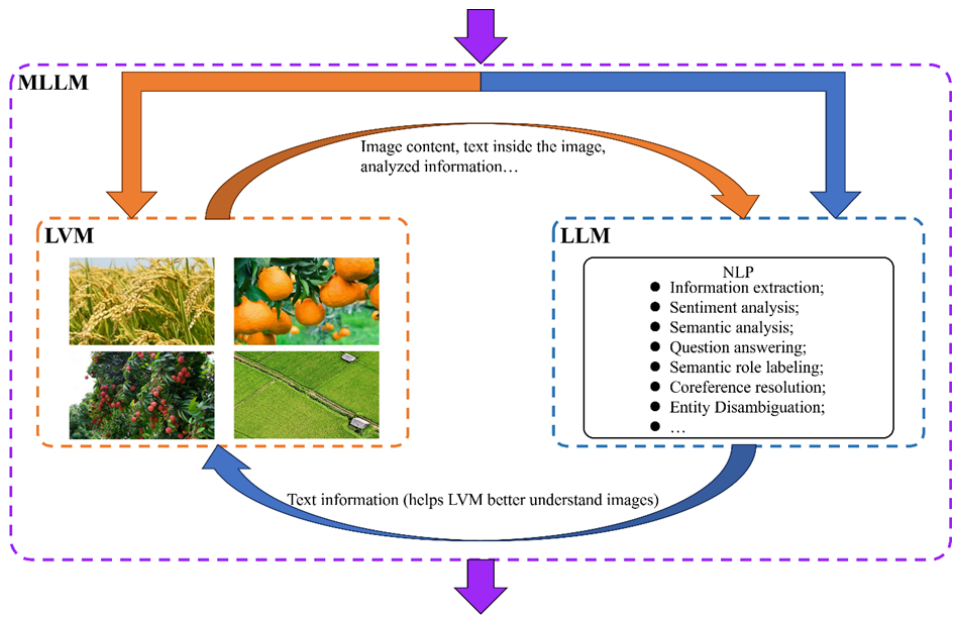

**Fig 4.** A multimodal large language model obtained by combining a large vision model and a large language model.

Nevertheless, it is important to highlight that existing models primarily rely on text-image data and are mostly limited to QA tasks. Actually, there is a noticeable lack of applications in the realm of agricultural robotics that incorporate inputs like images, text, voice (Human instructions), and depth information (From LiDAR or laser sensors). These agricultural robots, commonly deployed for tasks such as fruit picking or crop monitoring [123], present a significant opportunity for the integration of multimodal data sources to enhance their capabilities. In short, large models lacking a high degree of multimodality perform fewer tasks and lack good applicability.

*4.2. Results and impact assessment*

For the evaluation of a large model, questions can be asked and the answer results of the model can be tested. According to ethics and timeliness (Fig 5), it can be divided into A: the results have timeliness and do not violate ethics; B: the results have timeliness but violate ethics; C: outdated results that do not violate ethics; D: outdated results that violate ethics. As time passes, A will gradually become C, but it does not mean that the C is useless. Also affected by time, B will gradually become D, and D is the worst-case evaluation, both unethical and outdated. There are individual differences in the evaluation of A and B, and different questioners have different opinions on whether the same answer violates ethics. Therefore, judgments should be made based on the questioner's own religious beliefs, race, and other reasons. In summary, from a priority perspective, A>C>B>D.

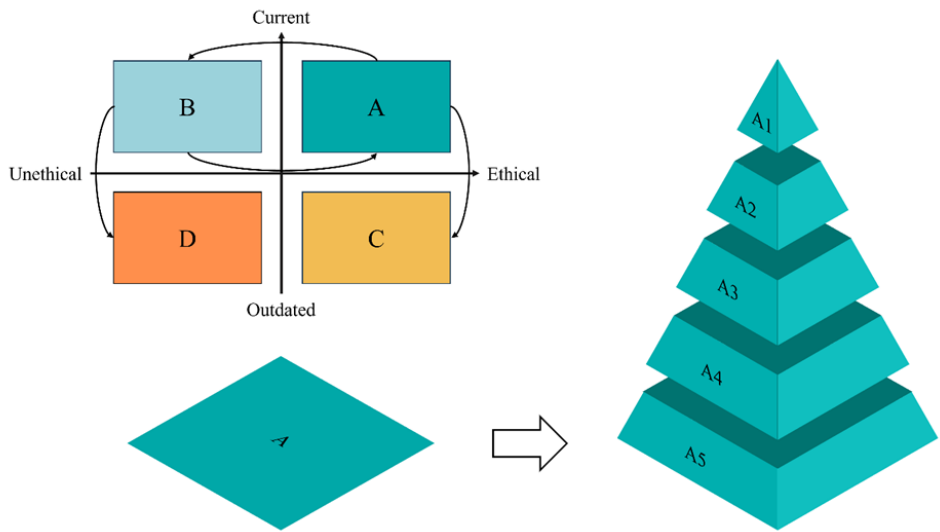

**Fig 5.** Impact assessment. Divide the criteria for evaluating large models into four categories: A, B, C, and D, and five subcategories: A1, A2, A3, A4, and A5.

Li et al. used a method to evaluate LVLM, namely polling-based object probing evaluation (POPE) [124]. Specifically, it is based on asking if there is any information in a picture and asking LVLM to answer "yes" or "no", e.g., "Is there a chair in the image?". Referring to POPE, researchers can formulate many questions and evaluate the model based on the accuracy of the answers. The formula is as follows:

$$A1: 80\% \leq Acc \leq 100\%$$
$$A2: 60\% \leq Acc < 80\%$$
$$A3: 40\% \leq Acc < 60\%$$
$$A4: 20\% \leq Acc < 40\%$$
$$A5: 0\% \leq Acc < 20\%$$

Where Acc represents accuracy. From a priority perspective, A1>A2>A3>A4>A5.

In a word, large models exhibit certain limitations in many tasks, and their reasoning results are often unethical or outdated. The model evaluation method proposed in this chapter refers to the research in [124] [125] [126], which relies more on manual evaluation and is only used as a reference. We hope to have large models that can conduct self-assessment in the future, reducing manual intervention.

## 5. Ethical issues and responsible use of large vision and language models in Agriculture

The performance of large models in agriculture demonstrates their superiority, and the potential of large models becomes evident when employed to enable predictive understanding of intricate systems. However, there are often ethical and responsibility issues in the development and deployment process of AI today. The digital gap between those who have the resources to develop and utilize large models and those who cannot afford to do so creates an inequality in accessing large models, resulting in an unfair distribution of risks and benefits [126]. Not only that, this divide is exacerbated by the presence of AI biases [127] [128]. Accordingly, in order to ensure ethical issues and responsible use of large models, this chapter starts from the ethical and responsibility issues in the agricultural large models and explore corresponding measures.

*5.1. Ethical issue of large language models to guide farming*

Predicting and solving ethics problems of large models in agriculture is a critical scientific and societal challenge. Although large models point the way for the future of smart agriculture, due to their characteristic of being influenced by close association, large models often learn some bad knowledge in addition to useful knowledge. Ethical issues have always been an indispensable topic of discussion in the process of technological progress (Such as the ethical issues discussed by Holmes et al. in the field of education regarding educational AI [129]), and we also need to pay attention to ethics issues when using large models in the agricultural direction. As mentioned below, many relevant institutions and personnel have put forward their own ideas on ethical issues related to large models.

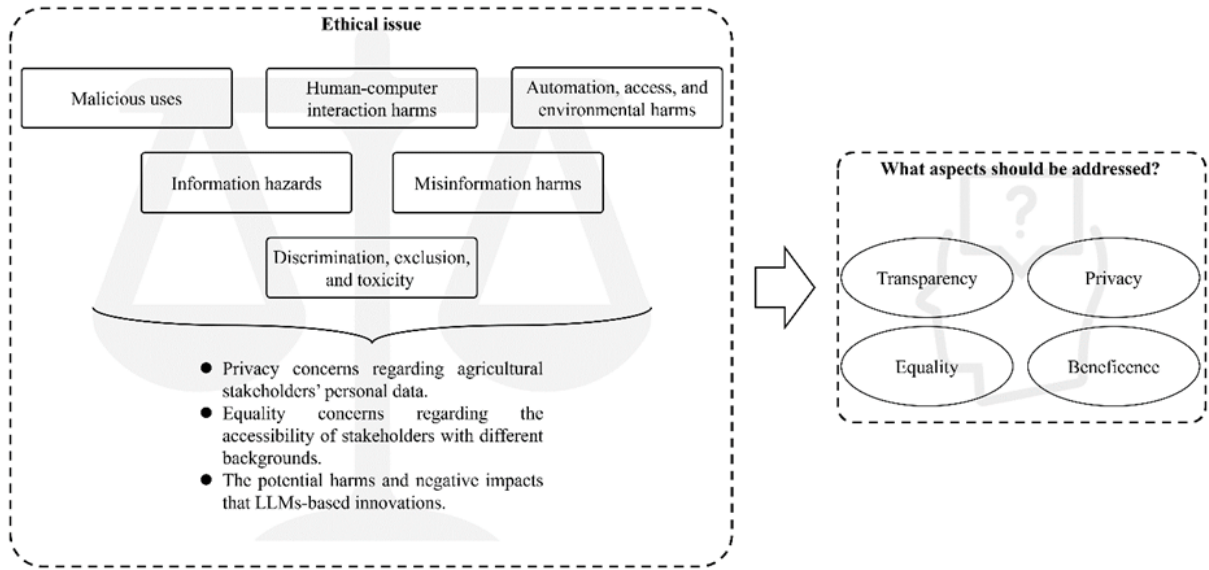

**Fig 6.** The ethical issues faced by large models.

Weidinger et al. [130] put forward six types of ethical risks(Fig 6): 1) Malicious uses, 2) Human-computer interaction harms, 3) Automation, access, and environmental harms, 4) Information hazards, 5) Misinformation harms, and 6) Discrimination, exclusion, and toxicity. Understanding these issues can help us responsibly use large models in the agricultural field.

- Malicious uses. Prior to the release of GPT-4, OpenAI hired a team of 50 experts and scholars to conduct a six-month adversarial test on GPT-4. Andrew White, a professor of chemical engineering at the University of Rochester who was invited to participate in the test, stated that early versions of GPT-4 could assist in the manufacture of chemical weapons and even provide a convenient manufacturing location. From the perspective of the agricultural sector, if this issue is not properly addressed, some may use large models to learn ways to destroy other people's farmland for the sake of profit, thereby allowing themselves to have a larger market. Over time, this will lead to vicious competition in the market.

- Human-computer interaction harms. The potential harms of human-computer interaction arise when users excessively trust a large model or mistakenly treat it as human.

- Automation, access, and environmental harms. The large model can give rise to automation, access, and environmental harms due to its potential environmental or downstream economic impacts.

- Information hazards. Due to the involvement of information from different countries, religions, and ethnicities, model outputs leaking or inferring sensitive information often led to political violence.

- Misinformation harms. A study discussed the potential risks of using poorly performing large models. The original intention of this study was to provide a natural language generation model in MOOC to respond to students and improve their participation rate [131]. Even so, due to the poor performance of the model, the corresponding negative results further reduced the enthusiasm of students. If a poorly performing large model is used in the agricultural field, it may mislead farmers in their judgment (Such as analysing incorrect disease types), not only causing further damage to crops in the farmland, but also making farmers increasingly distrust the large model. For this phenomenon, Angelone et al. proposed that warning labels can be applied to the content generated by the large model [132], but this also involves the trust issue of the large model in its own generated results.

- Discrimination, exclusion, and toxicity. Two researches have indicated that potential discrimination, exclusion, and toxicity issues may occur if adopting a model that is accurate but unfair [133] [134].

Despite Weidinger et al.'s viewpoint can provide us with a fundamental understanding of the risks associated with large models, manners of systematic ethical supervision of large models' research and innovation (R&I) are especially restricted. Coincidentally, the European Commission has officially approved comprehensive "ethics guidelines for trustworthy AI" specifically designed for R&I. These guidelines require that principal investigators recognize and tackle the ethical matters raised by their proposed research. Principal investigators are also required to adhere to ethical principles and relevant legislation in their work. In a similar vein, Stanford University's Ethics and Society Review necessitates researchers to distinguish potential societal hazards associated with their research and incorporate mitigation measures into their research design [135].

Furthermore, projects with large models have a vast amount of data and often raise ethics issues. Plant science data, for example, is not within the scope of the European Union General Data Protection Regulation (GDPR), creating challenges concerning data ownership. The issue of authorizing data sharing also varies across jurisdictions, adding to the complexity of the situation

[126]. Thus, relevant guidelines must take into account code of conduct for data sharing, privacy protection, and the overall governance of datasets.

*5.2. Responsible use in agriculture*

With the expanding development and utilization of large models, there is undoubtedly a growing need for agile and effective regulatory oversight. To address this issue, it may be necessary to use AI technology to assist in overseeing the development and deployment of large models. Regarding this aspect, the AI Act, which has been jointly agreed upon by the European Parliament and the Council of Europe, represents the first comprehensive set of harmonized rules on a global scale. It promotes responsible large model design and development by regulating large model across various applications and contexts based on a risk-based framework. Within the framework, careful consideration must be given to the level of risk involved and how to evaluate different large models as risk-free or low-risk.

To evaluate the risk level of a large model, we focus on four aspects: transparency, privacy, equality, and beneficence. On the other hand, in addition to developing and adhere to a strong regulatory framework that guides the development, deployment, and use of large models, regulatory methods also need to be considered. Consider the potential societal impact, potential harms, and long-term implications of the technology. Firstly, due to the wide applicability of large models, we cannot make a one size fits all approach. Regulation must adapt to specific issues in different domains. The United States' food and drug administration (FDA) has tailored potential regulatory methods for AI and machine learning technologies used in medical devices, categorizing them into three major categories based on risk levels: Class I (Low risk), Class II (Moderate risk), and Class III (High risk). Large models in agriculture can also be regulated according to the FDA's approach, dividing them into several types of models ranging from low risk to high risk. For example, genetically modified crops may have environmental impacts, food safety issues, and ecosystem damage, so large models targeting genetically modified crops should be included in high-risk types. For large models of ordinary crops, they can be classified as low-risk types. And the regulatory methods proposed by relevant departments should be made public to ensure transparency of information. Regulators can promote fairness in the deployment of agricultural large models by enforcing the use of diverse and representative data sources, which helps mitigate potential biases present in the training data [136].

From the perspective of beneficence and privacy, privacy issues related to large models have received little attention or investigation in reviewed research [137]. Specifically, if the training set used to train a large model contains some personal privacy information that has not been authorized by the information owner. The disregard for privacy concerns is especially worrisome considering that LLM-based innovations involve stakeholders' natural languages, which may contain personal and sensitive information pertaining to their private lives and identities [138]. If users unintentionally learn about this information while using a large model, it may cause harm to the beneficence of the information owner. Developers of large models should ensure they gain explicit consent from individuals before collecting and utilizing these personal data. Clearly communicate the purpose and scope of data usage, and offer individuals with the choice to choose out or request data deletion. Besides, limit the amount of personal and sensitive data collected and stored. Follow the principle of data minimization, ensuring that only necessary data is collected and retained. Anonymize or aggregate data whenever possible to protect individual privacy.

In general, governance approaches that promote responsible utilization of large models and focus on the outcomes rather than the technology itself will enhance research efforts and drive more innovation. By combining governance and ethics, we can harness the powerful synergy to expedite the implementation of large models in agriculture and other domains, fostering innovation at a larger scale.

## 6. Challenges and Future Directions

Although large models can play a powerful role in the field of agriculture, they still face challenges in many aspects.

*6.1. Technical and practical challenges*

1) Difficulty in obtaining agricultural data

Developing a large model suitable for agriculture requires a lot of relevant agricultural data, and there are many difficulties in the process of collecting these data. Due to factors such as crop types, growth stages, soil conditions, weather patterns, and agricultural practices, the complexity and high diversity of datasets make its collection and standardization a challenging task [10]. First of all, the collection of these data must be comprehensive and accurate, otherwise it cannot be guaranteed that the trained model will have good performance. Obtaining high-quality data is time-consuming, labour-intensive, and costly, particularly considering the requirement for ground truth labelling in supervised learning [139]. Secondly, as mentioned in Chapter 5, private data can pose risks to the use of large models. But to ensure the accuracy of the large model, some private data is necessary. However, as farmland is typically privately owned, farmers may exhibit hesitancy in sharing data due to concerns over privacy or latent mercantile exploitation. Thirdly, crops have a growth cycle, their growth process changes with the passage of time, influenced by daily fluctuations, seasonal changes, or annual variations. This requires the collection of time-series data, which introduces the other layer of complexity for data collection [140].

2) Low training efficiency

Training large models for agriculture applications presents significant challenges, especially when it comes to lengthy training times and significant costs, such as thousands of GPU hours and millions of training data are required [140]. Large models typically have a significantly higher number of parameters, which increases the computational demand during training. The computations involved in forward and backward passes through the layers of the model become more complex and time-consuming. Collecting, preprocessing, and loading massive data is time-consuming and resource-intensive.

3) Distribution shift

The problem of distribution shift is a major challenge when using large models in agriculture. When the data encountered by the model during deployment is obviously different from the data used in its training phase, a distribution shift will occur. The environmental conditions for collecting data may vary greatly in different regions and climates. These changes may include differences in crop types, soil conditions, weather patterns, and agricultural practices, all of which can lead to significant changes in data distribution [141]. The distribution shift will result in the trained large model not having strong applicability and may not achieve good results in some agricultural tasks. For example, it has been proven that applying large models directly to leaf segmentation tasks in a zero-shot means led to unsatisfactory performance, which can be attributed to possible distribution shifts [11].

4) The lag of data

After the trained large model is put into use, the data used for training has a certain timeliness for a long period of time. But after a long time, some data lags in time, and the results obtained by using a large model may deviate from the current facts (Fig 7).

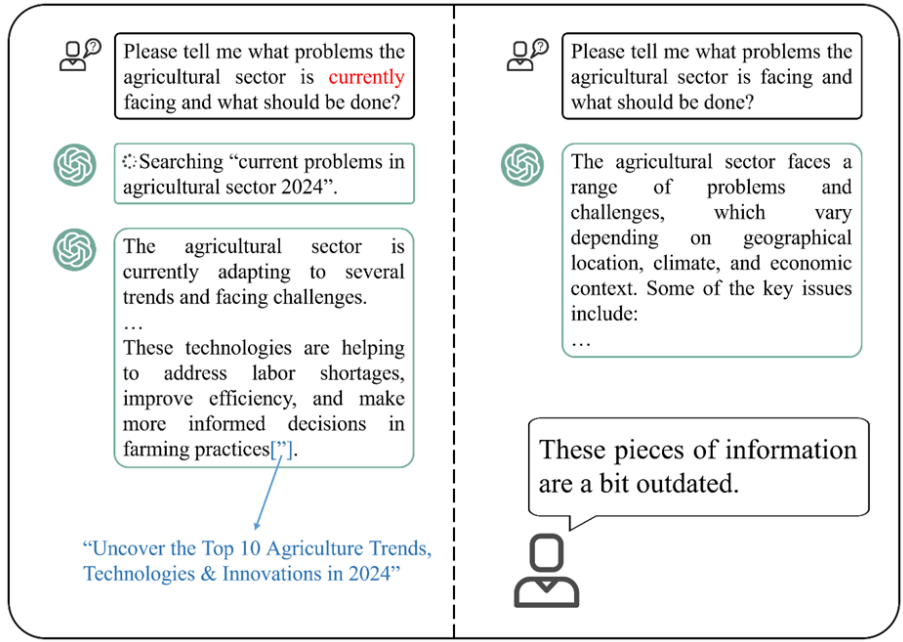

**Fig 7.** The lag of data.

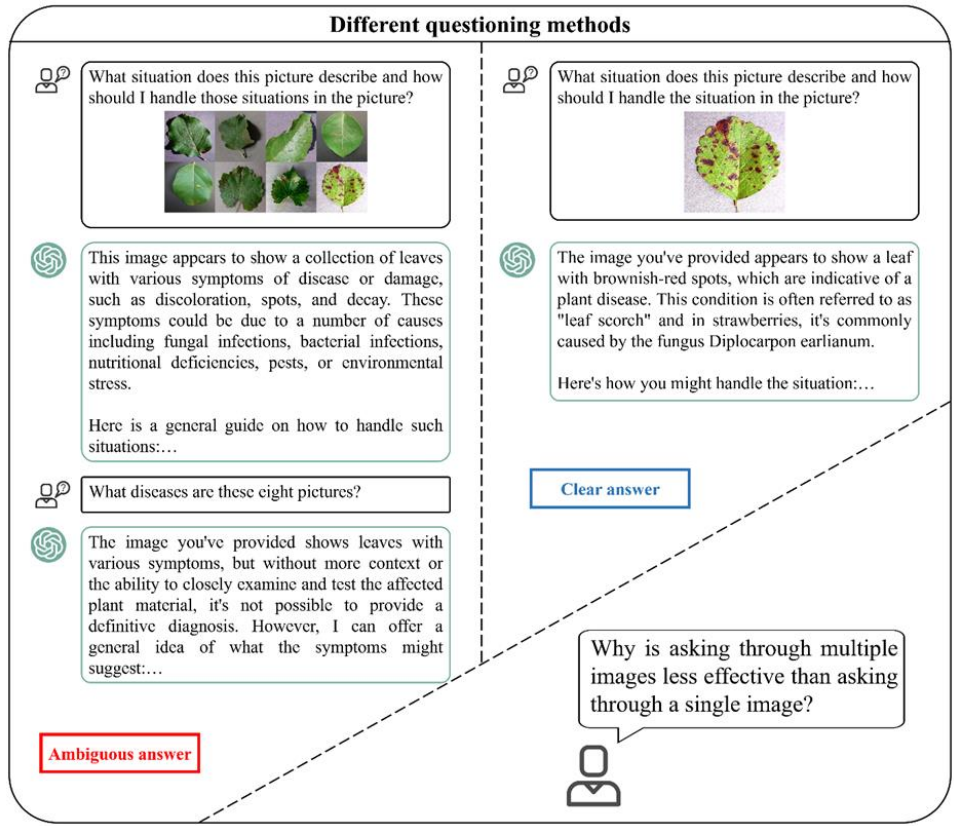

**Fig 8.** Different questioning methods can lead to different results.

5) Different methods, different results

When asking questions in different ways, the results obtained will also vary. Like Fig 8, when multiple images are spliced together for questioning, GPT-4 provides ambiguous answers; When only asking for one image, GPT-4 provides a clearer answer.

To clear these obstacles, future research and development work needs to pay attention to model optimization techniques such as model compression and efficient network structure design, reducing model size without affecting performance [142]. It is also necessary to provide update and maintenance functions for the model to ensure its timeliness. Developers need to write relevant usage instructions to help users get started quickly.

*6.2. Future trends in the integration of agricultural and food sectors and large models*

In the future, there will undoubtedly be agricultural large models with better performance and higher applicability. And the large models in agriculture should not be limited to text and image inputs. We believe that future multimodal agricultural models can support multimodal information such as videos (Analysing crops in videos) and audio (Tapping watermelons, and judging maturity through the sound emitted). On the other hand, agriculture is closely related to food, and the development of large models in agriculture is likely to promote the development of large models in the food domain. Trust is indispensable for agriculture and food system technologies given food's universality and importance to people [143]. Researchers need to navigate complicated social, political, economic, and environmental landscapes to develop appropriate large models in the food industry. In the future food industry, researchers will strive to establish trust with governmental agencies and funders, as well as with food system partners, to provide food and products that the public trusts [144].

Overall, although the agricultural large model still faces many challenges at present, we believe that through the joint efforts of relevant researchers in the future, these challenges can be properly addressed. And due to the close relationship between the food and agricultural domains, with the gradual development of agricultural large models, food large models will also receive further research, thereby achieving mutual positive feedback between the development of large models in these two fields.

## 7. Conclusion

In summary, this study investigated the application status of large models in the agricultural field. The current large models research in agricultural domain can bring great changes to agriculture, not only greatly improving the efficiency of agricultural production, but also further moving agriculture towards "smart agriculture" and "unmanned agriculture". However, there are still some unresolved issues in the development and deployment of agricultural large models. Although some studies have proposed solutions to these issues, further research is needed to ensure the applicability and reliability of large models in the field of agriculture. The food field is closely related to the agricultural field, and in the process of developing the agricultural large models, the food large models will also be further developed. These two different fields of large models will interact with each other, forming positive feedback. In the future, research on agricultural large models still needs to be further improved in terms of applicability and reliability, to prevent erroneous information from misleading farmers and causing damage to farmland, as well as farmers' distrust of new technologies. We hope this paper can serve as a support and cornerstone for the development of future agricultural models.

# Acknowledgements

**Funding:** Our research was supported by the Natural Science Foundation of Guangxi (No. 2024 GXNSFBA010381), the Guangxi Science and Technology Base and Talent Project (Guike AD21075018), and the National Natural Science Foundation of China (No. 62361006).

**Author contributions:** The two authors (Hongyan Zhu and Shuai Qin) contributed equally to this study and shared the first authorship. **Hongyan Zhu:** Funding acquisition, Methodology, Supervision, Writing - review & editing. **Shuai Qin:** Conceptualization, Formal analysis, Investigation, Methodology, Resources, Software, Writing - Original draft, Writing - review & editing, Visualization. **Min Su:** Funding acquisition, Writing - review & editing. **Chengzhi Lin:** Software, Writing - Original draft. **Anjie Li:** Software. **Junfeng Gao:** Conceptualization, Investigation, Project administration, Writing - review & editing. All authors have read and agreed to the published version of the manuscript.

**Competing interests:** The authors declare that they have no known competing financial interests or personal relationships that could have appeared to influence the work reported in this paper.